\documentclass[preprint,12pt,3p]{elsarticle}
\usepackage{amssymb}
\usepackage{amsmath,bm}
\usepackage{hyperref}
\usepackage{url}
\usepackage[utf8]{inputenc}
\usepackage{makecell}
\usepackage{dblfloatfix}
\newcolumntype{P}[1]{>{\centering\arraybackslash}p{#1}}
\newcolumntype{M}[1]{>{\centering\arraybackslash}m{#1}}
\usepackage{pifont}% http://ctan.org/pkg/pifont
\newcommand{\cmark}{\ding{51}}%
\usepackage{xcolor,pifont}
\newcommand*\colourcheck[1]{\expandafter\newcommand\csname #1check\endcsname{\textcolor{#1}{\ding{52}}}}
\usepackage{tikz}
\usetikzlibrary{automata,positioning}
\usepackage[font=footnotesize]{caption}

\journal{Pattern Recognition Letters}
\let\oldhref\href
\renewcommand{\href}[2]{\oldhref{#1}{\hbox{#2}}}
\journal{Image and Vision Computing}

\begin{document}

\begin{frontmatter}

\title{An Attention-Based Deep Learning Model for Multiple Pedestrian Attributes Recognition}

\author[label1,label2]{Ehsan Yaghoubi\corref{cor1}}
\address[label1]{IT: Instituto de Telecomunicações}
\address[label2]{University of Beira Interior}

\cortext[cor1]{Ehsan.Yaghoubi@ubi.pt}
%\fntext[label3]{I also want to inform about\ldots}
%\fntext[label4]{Small city}

% \ead{author.one@mail.com}
% \ead[url]{author-one-homepage.com}

\author[label3]{Diana Borza}
\address[label3]{Technical University of Cluj-Napoca}
%\ead{author.two@mail.com}

\author[label14]{João Neves}
\address[label4]{TomiWorld}

\author[label3]{Aruna Kumar}

\author[label1,label2]{Hugo Proença}

\begin{abstract}
The automatic characterization of pedestrians in surveillance footage is a tough challenge, particularly when the data is extremely diverse with cluttered backgrounds, and subjects are captured from varying distances, under multiple poses, with partial occlusion. Having observed that the state-of-the-art performance is still unsatisfactory, this paper provides a novel solution to the problem, with two-fold contributions: 1) considering the strong semantic correlation between the different full-body attributes, we propose a multi-task deep model that uses an element-wise multiplication layer to extract more comprehensive feature representations. In practice, this layer serves as a filter to remove irrelevant background features, and is particularly important to handle complex, cluttered data; and 2) we introduce a weighted-sum term to the loss function that not only relativizes the contribution of each task (kind of attributed) but also is crucial for performance improvement in multiple-attribute inference settings. Our experiments were performed on two well-known datasets (RAP and PETA) and point for the superiority of the proposed method with respect to the state-of-the-art. The code is available at \url{https://github.com/Ehsan-Yaghoubi/MAN-PAR-}.
\end{abstract}


\end{frontmatter}

\section{Introduction}
\label{sec1}

\begin{figure}
\centering
\includegraphics[scale=0.36]{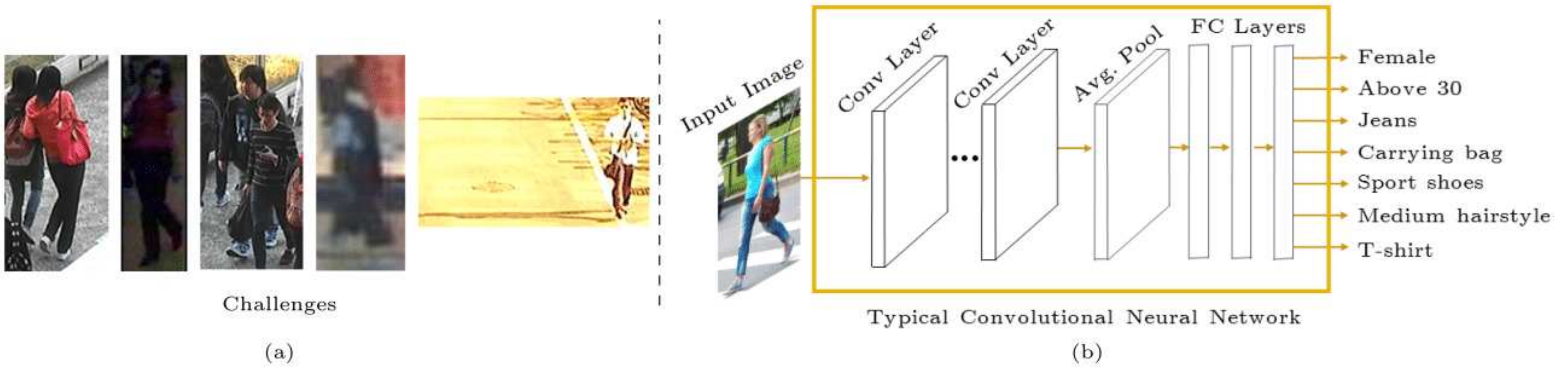}
\caption{ (a) Examples of some of the challenges in the PAR problem: crowded scenes, poor illumination conditions, and partial occlusions. (b) Typical structure of PAR networks, which receive a single image and perform labels inference.}
\label{fig: typicalCNN_}
% \vspace{-0.5 cm}
\end{figure}

The automated inference of pedestrian attributes is a long-lasting goal in video surveillance and has been the scope of various research works \citep{mabrouk2018abnormal}  \citep{kumari2015facial}. Commonly known as \emph{pedestrian attribute recognition} (PAR), this topic is still regarded as an open problem, due to extremely challenging variability factors such as occlusions, viewpoint variations, low-illumination, and low-resolution data (Fig. \ref{fig: typicalCNN_} (a)).

Deep learning frameworks have repeatedly been improving the state-of-the-art in many computer vision tasks, such as object detection and classification, action recognition and soft biometrics inference. In the PAR context, several models have been also proposed \citep{schmidhuber2015deep}, \citep{liu2017survey}, with most of these techniques facing particular difficulties to handle the heterogeneity of visual surveillance environments.

Researchers have been approaching the PAR problem from different perspectives \citep{wang2019pedestrian}: \citep{li2015multi}, \citep{sudowe2015person}, \citep{abdulnabi2015multi} proposed deep learning models based on \emph{full-body} images to address the data variation issues, while \citep{liu2018localization}, \citep{gkioxari2015actions}, \citep{li2016human}, \citep{chen2018pedestrian} described \emph{body-part} deep learning networks to consider the fine-grained features of the human body parts. Other works focused particularly on the \emph{attention mechanism} \citep{sarafianos2018deep}, \citep{sarfraz2017deep}, \citep{li2016human}, and typically performed additional operations in the output of the mid-level and high-level convolutional layers. However, learning a comprehensive feature representation of pedestrian data, as the backbone for all those approaches, still poses some challenges, mostly resulting from the multi-label and multi-task intrinsic properties of PAR networks.

In opposition to previous works that attempted to jointly extract local, global and fine-grained features from the \emph{input image}, in this paper, we propose a multi-task network that processes the \emph{feature maps} and not only considers the correlation among the attributes but also captures the foreground features using a hard attention mechanism. The attention mechanism yields from the element-wise multiplication between the feature maps and a foreground mask that is included as a layer on top of the backbone feature extractor. Furthermore, we describe a weighted binary cross-entropy loss, where the weights are determined based on the number of categories (e.g., gender, ethnicity, age, \ldots) in each task. Intuitively, these weights control the contribution of each category during training and are the key to avoid the predominance of some labels over the others, which was one of the problems we identified in our evaluation of the previous works. In the empirical validation of the proposed method, we used two well-known PAR datasets (PETA and RAP) and three baseline methods considered to represent the state-of-the-art.

\begin{figure}
\centering
\begin{tikzpicture}
\node[anchor=center] (p3) at (0,0) {\includegraphics[scale=.30]{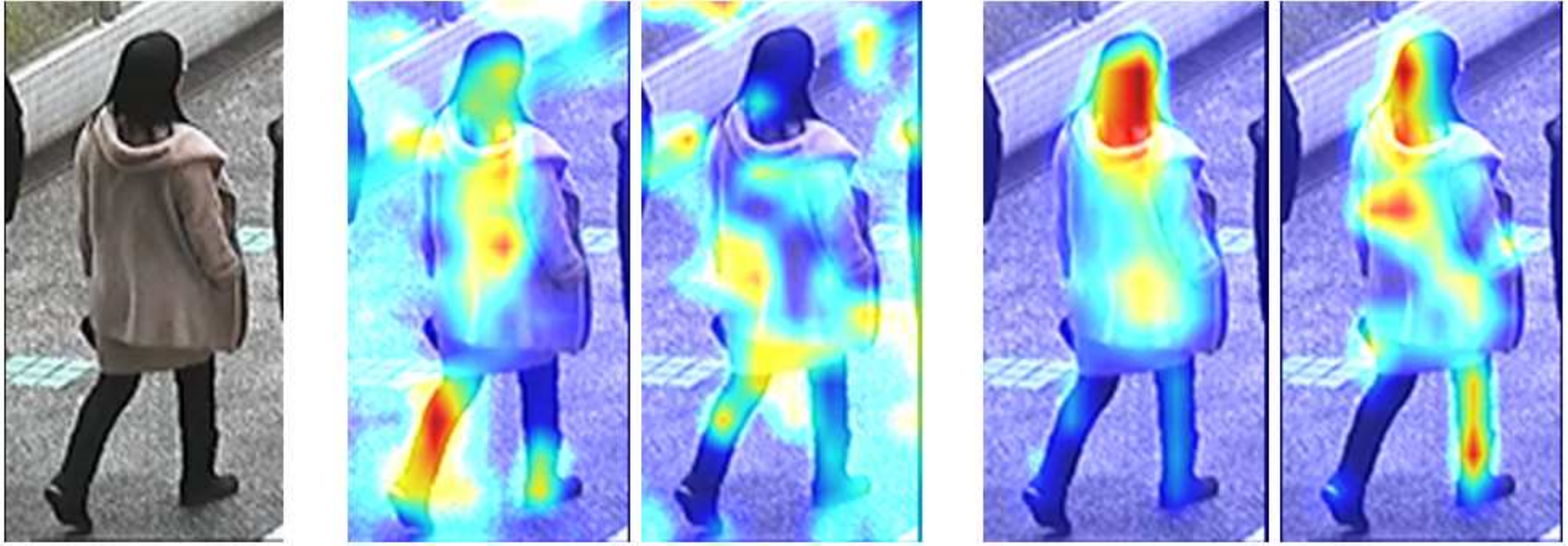}};
\node[align=center] at (-4.2,-2.1) {\tiny Original};
\node[align=center] at (-1,-2.1) {\tiny Typical methods};
\node[align=center] at (3.2,-2.1) {\tiny Proposed method with different settings};
\end{tikzpicture}
\caption{Comparison between the attentive regions obtained typically by previous methods \citep{li2018richly}, \citep{zhu2017multi} and ours solution, while inferring the \emph{Gender} attribute. Note the less importance given to background regions by our solution with respect to previous techniques.}
\label{fig: compare}
% \vspace{-0.5 cm}
\end{figure}
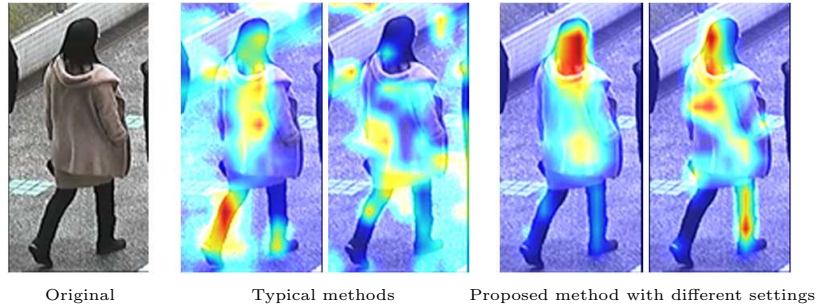

The contributions of this work can be summarized as follows:
\begin{enumerate}
  \item We propose a multi-task classification model for PAR that its main feature is to focus on the foreground (human body) features, attenuating the effect of background regions in the feature representations (Fig. \ref{fig: compare});
  \item We describe a weighted sum loss function that effectively handles the contribution of each category (e.g., gender, body figure, age, etc.) in the optimization mechanism, which prohibits some of the categories to predominate over the others during the inference step;
  \item Inspired by the attention mechanism, we implement an element-wise multiplication layer that simulates hard attention in the output of the convolutional layers, which particularly improves the robustness of feature representations in highly heterogeneous data acquisition environments.
\end{enumerate}

The remainder of this paper is organized as follows: Section \ref{sec:Related Works} summarises the PAR-related literature, and section \ref{sec:Proposed Method} describes our method. In section \ref{sec:Experimental Details}, we provide the empirical validation details and discuss the obtained results. Finally, conclusions are provided in section \ref{sec: Conclusion}.

\section{Related Work}
\label{sec:Related Works}
The ubiquity of CCTV cameras has been rising the ambition of obtaining reliable solutions for the automated inference of pedestrian attributes, which can be particularly hard in case of crowded urban environments. Given that face close-shots are rarely available at far distances, PAR upon full-body data is of practical interest. In this context, the earlier PAR methods focused individually on a single attribute and used handcrafted feature sets to feed classifiers such as SVM or AdaBoost\citep{PARatFAR}, \citep{zhu2013pedestrian} \citep{layne2014attributes}. More recently, most of the proposed methods were based on deep learning frameworks, and have been repeatedly advancing the state-of-the-art performance \citep{tan2019attention}, \citep{li2019visual}, \citep{zhao2019recurrent}, \citep{lou2019mse}.

In the context of deep learning, \citep{zhu2015multi} proposed a multi-label model composed of several CNNs working in parallel, and specialized in segments of the input data. \citep{li2015multi} compared the performance of single-label versus multi-label models, concluding that the semantic correlation between the attributes contributes to improve the results. \citep{sudowe2015person} proposed a parameter sharing scheme over independently trained models. Subsequently, inspired by the success of Recurrent Neural Networks, \citep{wang2017attribute} proposed a Long Short-Term Memory (LSTM) based model to learn the correlation between the attributes in low-quality pedestrian images. Other works also considered information about the subjects pose \citep{li2018pose}, body-parts \citep{yang2016attribute} and, viewpoint \citep{liu2018localization}, \citep{sarfraz2017deep}, claiming to improve performance by obtaining better feature representations. In this context, by aggregating multiple feature maps from low, mid, and high-level layers of the CNN, \citep{Liu_2017_ICCV} enriched the obtained feature representation. For a comprehensive overview of the existing human attribute recognition approaches, we refer the readers to \citep{wang2019pedestrian}.

\section{Proposed Method}
\label{sec:Proposed Method}

 As illustrated in Fig. \ref{fig: compare}, our primary motivation is to provide a PAR pipeline that is robust to background-based irrelevant features, which should contribute for improvements in performance, particularly in crowded scenes that partial occlusions of human body silhouettes occur (Fig. \ref{fig: typicalCNN_} (a) and Fig. \ref{fig: compare}).

 \subsection{Overall Architecture}
 \label{sec: Architecture of the Proposed Network}
\begin{figure*}[t]
\centering
\includegraphics[scale=.355]{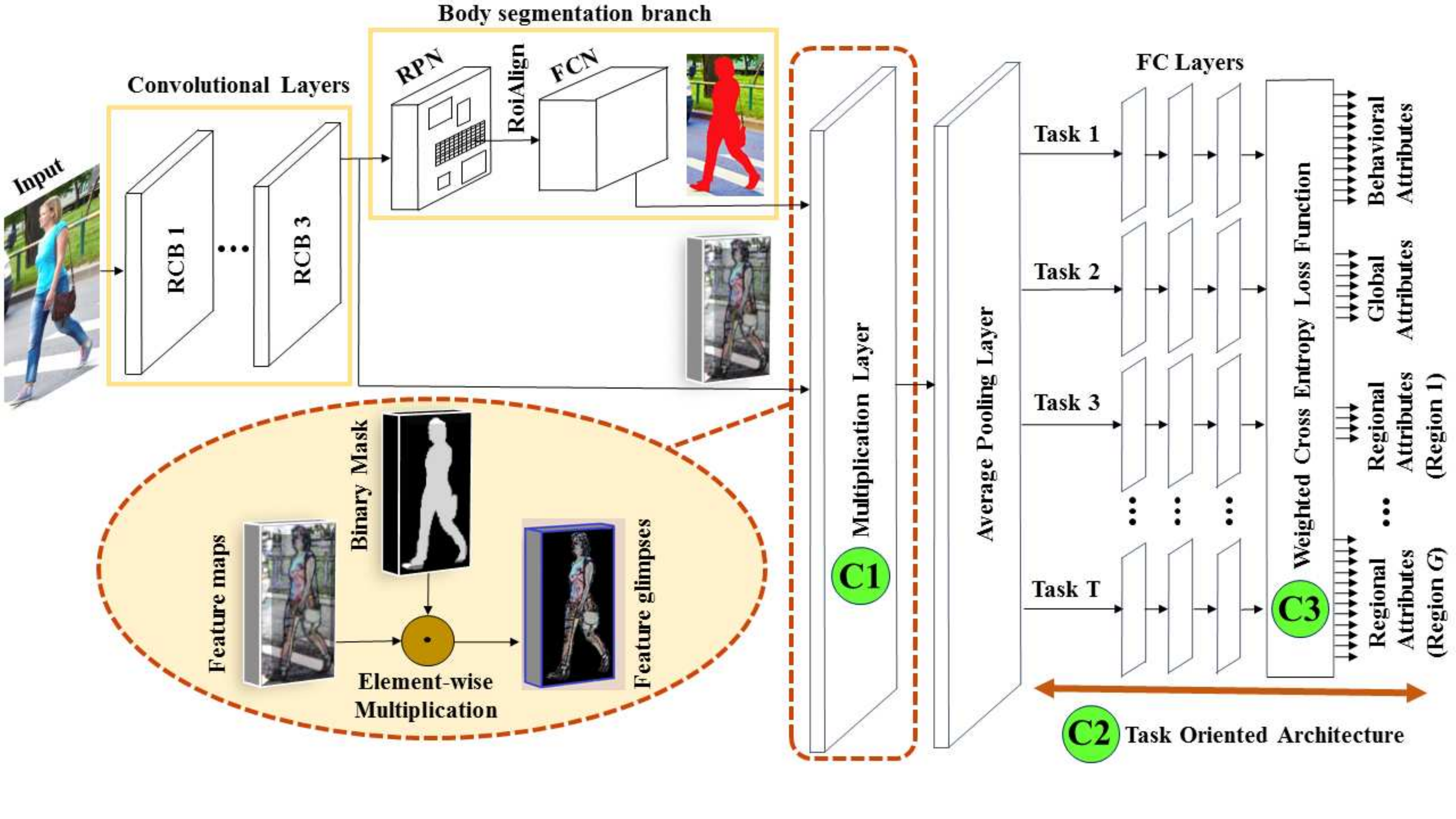}
\caption{Overview of the major contributions (C$i$) in this paper. C1) the element-wise multiplication layer receives a set of feature maps $\bm{F}_{H\times W\times D}$ and a binary mask $\bm{M}_{H\times W\times D}$, and outputs a set of \emph{attention glimpses}. C2) The multitask-oriented architecture provides to the network the ability to focus on the local (e.g., head accessories, types of shoes), behavioral (e.g., talking, pushing), and global (e.g., age, gender) features (visual results are given in Fig.  \ref{fig:body_part_heatmap}). C3) a weighted cross-entropy loss function not only considers the interconnection between the different attributes, but also handles the contribution of each label in the inference step. RCB is the abbreviation for Residual Convolutional Block, illustrated in Fig. \ref{fig: residual}. RPN, FCN, and FCL stand for Region Proposal Network, Fully Connected Network, and Fully Connected layer, respectively.}
\label{fig: Network}
% \vspace{-0.3 cm}
\end{figure*}

 Fig. \ref{fig: Network} provides an overview of the proposed model, inferring the complete set of attributes of a pedestrian at once, in a single-shot paradigm. Our pipeline is composed of four main stages: 1) the \emph{convolutional layers}, as general feature extractors; 2) the \emph{body segmentation module}, that is responsible for discriminating between the foreground/background regions; 3) the \emph{multiplication layer}, that in practice implements the attention mechanism; and 4) the \emph{task-oriented branches}, that avoid the predominance of some of the labels over others in the inference step.

 At first, the input image feeds a set of convolutional layers, where the local and global features are extracted. Next, we use the body segmentation module to obtain the binary mask of the pedestrian body. This mask is used to remove the background features, by an element-wise multiplication with the feature maps. The resulting features (that are free of background noise) are then compressed using an average pooling strategy. Finally, for each \emph{task}, we add different fully connected layers on top of the network, not only to leverage the useful information from other tasks but also to improve the generalization performance of the network. We have adopted a multi-task network, because the shared convolutional layers extract the \emph{common} local and global features that are necessary for all the tasks (i.e., behavioral attributes, regional attributes, and global attributes) and then, there are separate branches that allow the network to focus on the most important features for each task.

\subsection{Convolutional Building Blocks}
The implemented convolution layers are based on the concept of residual block. Considering $\bm{x}$ as the input of a conventional neural network, we want to learn the true distribution of the output $H (\bm{x}) $. Therefore, the difference (residual) between the input and output is $R (\bm{x}) =H (\bm{x}) -\bm{x}$, and can be rearranged to $H (\bm{x}) =R (\bm{x}) +\bm{x}$. In other words, traditional network layers learn the true output $H (\bm{x}) $, whereas residual network layers learn the residual $R (\bm{x}) $. It is worth mentioning that it is easier to learn the residual of the output and input, rather than only the true output \citep{he2016deep}. In fact, residual-based networks have the degree of freedom to train the layers in residual blocks or skip them. As the optimal number of layers depends on the complexity of the problem under study, adding skip connections makes the neural network active in training the useful layers.

\begin{figure}
\centering
\includegraphics[scale=.35]{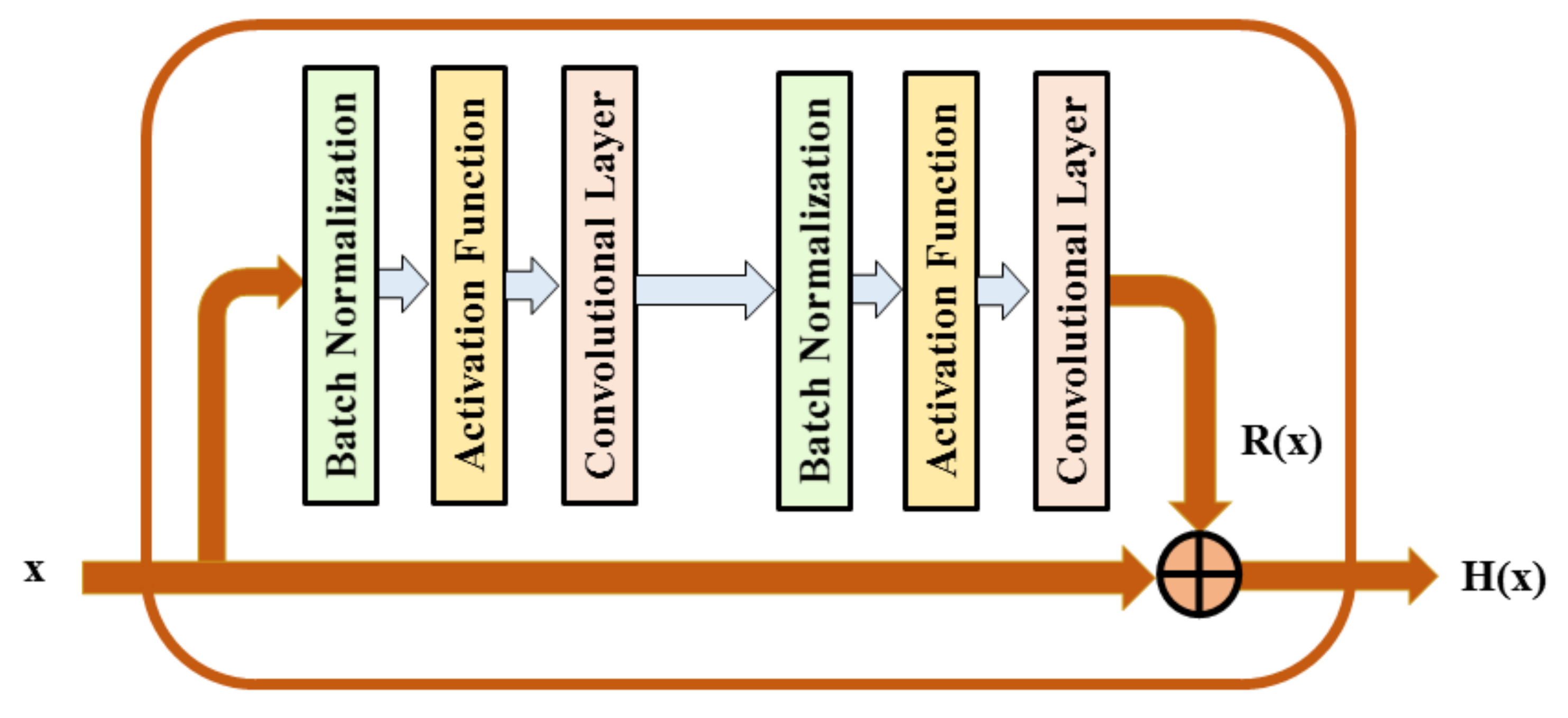}
\caption{Residual convolutional block in which the input $\bm{x}$ is considered a skip connection.}
\label{fig: residual}
% \vspace{-0.5 cm}
\end{figure}

 There are various types of residual blocks made of different arrangements of the Batch Normalization (BN) layer, activation function, and convolutional layers. Based on the analysis provided in \citep{he2016identity}, the forward and backward signals can directly propagate between two blocks, and optimal results will be obtained when the input $\bm{x}$ is used as skip connection (Fig. \ref{fig: residual}).

  \subsection{Foreground Human Body Segmentation Module}
  We used the Mask R-CNN \citep{he2017mask} model to obtain the full-body human  masks. This method adopts a two-stage procedure after the convolutional layers: $i)$ a Region Proposal Network (RPN) \citep{ren2015faster} that provides several possibilities for the object bounding boxes, followed by an alignment layer; and $ii)$ a Fully Convolutional Network (FCN) \citep{long2015fully} that infers the bounding boxes, class probabilities, and the segmentation masks.
  %Here, we have chosen to use a complex network instead of a simple semantic segmentation CNN.

  \subsection{Hard Attention: Element-wise Multiplication Layer}
    %Foreground segmentation have always been at the center of attention as significantly affects the recognition rate in computer vision tasks. To study the effectiveness of the proposed attention network with feature map level segmentation, we aim to force the network to focus on the foreground features.
  %As discussed in the literature review section, most works perform the operations directly on the image: however, inspiring from \citep{girshick2015fast}, we suggest feature map-level operations due to lower computational complexity, higher speed, and intrinsic hard parameter sharing \citep{DBLP}.

  The idea of an attention mechanism is to provide the neural network with the ability to focus on a feature subset. Let $\bm{I}$ be an input image, $\bm{F}$ the corresponding feature maps, $\bm{M}$ an attention mask, $f_\phi(\bm{I})$ an attention network with parameters $\phi$, and $\bm{G}$ an attention glimpse (\emph{i.e.}, the result of applying an attention mechanism to the image $\bm{I}$). Typically, the attention mechanism is implemented as $F=f_\phi(\bm{I})$, and $\bm{G}=\bm{M} \odot \bm{F}$, where $\odot$ is an element-wise multiplication. In soft attention, features are multiplied with a mask of values between zero and one, while in the hard attention variant, values are binarized and - hence - they should be fully considered or completely disregarded.

  In this work, as we produce the foreground binary masks, we applied a hard attention mechanism on the output of the convolutional layers. To this end, we used an element-wise multiplication layer that receives a set of feature maps $\bm{F}_{H\times W\times D}$ and a binary mask $\bm{M}_{H\times W\times D}$, and returns a set of attention glimpses $\bm{G}_{H\times W\times D}$, in which $H$, $W$, and $D$ are the height, weight, and the number of the feature maps, respectively.

%in which after four stages of convolution, batch normalization, and activation layers we insert a multiplication layer. Forth stage output of the traditional ResNet50 and  The inputs of this lay

 \subsection{Multi-Task CNN Architecture and Weighted Loss Function}
 \label{subsec: Weighted Loss Function}
%Let $N$ be the number of pedestrian images labeled with $L$ attributes and each image is represented as $I_i, i \in \begin{Bmatrix}1, ..., N\end{Bmatrix}$ with the corresponding label $y_i$ such that each element of the label vector $y_i$ is defined as $y_{il}$, $l \in 1, ..., L$ and $y_{il} \in \begin{Bmatrix}0, 1\end{Bmatrix}$. If $y_{il}= 1$, sample $I_i$ has the $l$’th attribute otherwise it does not.

We consider multiple soft label \emph{categories} (e.g., gender, age, lower-body clothing, ethnicity and hairstyle), with each of these including two or more \emph{classes}. For example, the category of \emph{lower-body clothing} is composed of 6 classes: \{'pants', 'jeans', 'shorts', 'skirt', 'dress', 'leggings'\}. As stated above, there are evident semantic dependencies between most of the labels (e.g., it is not likely that someone uses a 'dress' and 'sandals' at the same time). Hence, to model these relations between the different categories, we use a hard parameter sharing strategy\citep{DBLP} in our multi-task residual architecture. Let $T$, $C_{t}$, $K_{c}$, $N_{k}$ be the number of tasks, the number of categories (labels) in each task, the number of classes in each category, and the number of samples in each class, respectively.

During the learning phase, the model $\mathcal{H}$ receives one input image $\bm{I}$, its binary mask $\bm{S}$, the ground truth labels $\bm{Y}$, and returns $\bm{\hat{Y}}$ as the predicted attributes (labels):

 \begin{align}
 \bm{\hat{Y}}= \begin{Bmatrix}&\hat{y}_{t,c_{t},k_{t}}|t\in \begin{Bmatrix}1,...,T\end{Bmatrix}, c\in \begin{Bmatrix}1,...,C_{t}\end{Bmatrix}, k\in \begin{Bmatrix}1,...,K_{c}\end{Bmatrix},\\
 &T, C_{t}, K_{c}\in \mathbb{N}, \hat{y}_i\in\begin{Bmatrix}1,0\end{Bmatrix} \end{Bmatrix},
 \label{equ: Net_output}
 \end{align}
in which $\bm{\hat{y}}_{t,c,k}$ denotes the predicted attributes.

The key concept of the learning process is the loss function. In the single attribute recognition\citep{liu2015deep} setting, if the $n$-th image $\bm{I}_n$, $(n = 1,..., N)$ is characterized by the $m$-th attribute, $(m = 1,..., M)$, then $\bm{y}_{nm}=1$; otherwise, $\bm{y}_{nm}=0$.  In case of having multiple attributes (multi-task), the predicting functions are in the form of $\Phi =\begin{Bmatrix} \Phi_1, \Phi_2, ..., \Phi_m, ... , \Phi_M \end{Bmatrix}$, and $\Phi_{m}(\bm{I{}'})\in \begin{Bmatrix} 1, 0 \end{Bmatrix}$. We define the minimization of the loss function over the training samples for the $m$th attribute as:
\begin{equation}
\Psi_{m}=argmin_{\Psi_{m}}\sum_{n=1}^{N} \mathcal{L} \Big(\Phi_{m} (I_n,\Psi_{m}), y_{nm}\Big),
\label{equ: separated_loss}
\end{equation}
where $\Psi_m$ contains a set of optimized parameters related to the $m$-th attribute, while $\Phi_{m} (I_n,\Psi_{m})$ returns the predicted label ($\bm{\hat{y}_{nm}}$) for the $m$-th attribute of the image $I_n$. Besides, $\mathcal{L}(.)$ is the loss function that measures the difference between the predictions and ground-truth labels.

Considering the interconnection between attributes, one can define a unified multi-attribute learning model for all the attributes. In this case, the loss function jointly considers all the attributes:
\begin{equation}
    \Psi=argmin_{\Psi}\sum_{m=1}^{M}\sum_{n=1}^{N} \mathcal{L} \Big(\Phi_{m} (I_n,\Psi_{m}), y_{nm}\Big),
    \label{equ: multi_att_loss}
\end{equation}
 in which $\Psi$ contains the set of optimized parameters related to all attributes.

In opposition to the above-mentioned functions, in order to consider the contribution of each category in the loss value, we define a weighted sum loss function:
 \begin{align}
    \Psi=argmin_{\Psi}
    \sum_{t=1}^{T}
    &\sum_{c=1}^{C_{t}}
    \sum_{k=1}^{K_{c}}
    \sum_{n=1}^{N_k}
    \frac{1}{\mathcal{R}_{c}} \mathcal{L} \Big(\Phi_{tck} (I_n,\Psi_{tck}), y_{tckn}\Big), \nonumber\\
    %\label{equ: multi_task_loss}
\end{align}
where $\quad {\mathcal{R}_{c}} \in \{R_1,...,R_{C_{t}}\}$ are scalar values corresponding to the number of classes in the categories $1,...,C_{t}$.

Using the \emph{sigmoid} activation function for all classes in each category, we can formulate the \emph{cross-entropy} loss function as:

\begin{align}
Loss=-
\sum_{t=1}^{T}
\sum_{c=1}^{C_{t}}
\sum_{k=1}^{K_{c}}
\sum_{n=1}^{N_k}\frac{1}{{n} \mathcal{R}_{c}} \Big(y_{tckn} log(\hat p_{tckn}) +(1-y_{tckn}) log(1-\hat p_{tckn})\Big),
%\label{equ: CE_loss}
\end{align}
where $\bm{y}_{tckn}$ is the binary value that relates the class label $k$ in category $c$. The ground-truth label for observation $n$ and $\bm{{\hat p}}_{tckn}$ is the predicted probability of the observation $n$.

\section{Experiments and Discussion}
\label{sec:Experimental Details}
The proposed PAR network was evaluated on two well-known datasets: the PETA \citep{PARatFAR} and the Richly Annotated Pedestrian (RAP) \citep{li2018richly}, with both being among the most frequently used benchmarks in PAR experiments.

\begin{table}[!t]
\scriptsize
    \centering
    \caption{RAP dataset annotations}
    \label{table: rap_annot}
    \begin{tabular}{||m{2.7cm} | m{8cm}||}
 \hline
 Branch & Annotations  \\ [0.5ex]
 \hline\hline
 Soft Biometrics & Gender, Age, Body figure, Hairstyle, Hair color  \\
 \hline
 Clothing Attributes & Hat, Upper body clothes style and color, Lower body clothes style and color, Shoe style \\
 \hline
 Accessories &  Glasses, Backpack, Bags, Box \\
 \hline
 Action & Telephoning, Talking, Pushing, Carrying, Holding, Gathering  \\
 \hline
\end{tabular}
% \vspace{-0.3 cm}
\end{table}

\subsection{Datasets}
RAP \citep{li2018richly} is the largest and the most recent dataset in the area of surveillance, pedestrian recognition, and human re-identification. It was collected at an indoor shopping mall with 25 HD cameras (spatial resolution $1,280 \times 720$) during one month. Benefiting from a \emph{motion detection and tracking algorithm}, authors have processed the collected videos, which resulted in 84,928 human full-body images. The resulting bounding boxes vary in size from $33 \times 81$ to $415 \times 583$. The annotations provide information about the viewpoint ('front', 'back', 'left-side', and 'right-side'), body occlusions, and body-part pose, along with a detailed specification of the train-validation-test partitions, person ID, and 111 binary human attributes. Due to the unbalanced distribution of the attributes and insufficient data for some of the classes, only 55 of these binary attributes were selected~\citep{li2018richly}. Table \ref{table: rap_annot} shows the categories of these attributes. It is worth mentioning that, as the annotation process is performed per subject instance, the same identity may have different attribute annotations in distinct samples.

PETA \citep{PARatFAR} contains ten different pedestrian image collections gathered in outdoor environments. It is  composed of 19,000 images corresponding to 8,705 individuals, each one annotated with 61 binary attributes, from which 35 were considered with enough samples and selected for the training phase. Camera angle, illumination, and the resolution of images are the particular variation factors in this set.

\subsection{Evaluation Metrics}
 PAR algorithms are typically evaluated based on the standard classification accuracy per attribute, and on the mean accuracy ($\overline{mA}$) of the attribute. Further, the mean accuracy over all attributes was also used \citep{he2017adaptively}, \citep{lin2019improving}:

\begin{equation}
     \overline{mA}=\frac{1}{2M}\sum_{m=1}^{M} \left(\frac{\mathcal{\hat P}_m}{\mathcal{P}_m}+\frac{\mathcal{\hat N}_m}{\mathcal{N}_m}\right),
\end{equation}
where $m$ denotes one attribute, and $M$ is the total number of attributes. For each attribute $m$, $\mathcal{P}_m$, $\mathcal{N}_m$, $\mathcal{\hat P}_m$, and $\mathcal{\hat N}_m$ stand for the number of positive samples, negative samples, correctly recognized as positive samples, correctly identified as negative samples.

% Also, according to a recommendation of~\citep{li2018richly} the accuracy ($\mathcal{A}$), precision ($\mathcal{P}$), recall ($\mathcal{R}$) rate, and ($\mathcal{F}$) value were also obtained:

% \begin{gather}
%      \mathcal{A}=\frac{1}{N} \sum_{n=1}^{N}\frac{\left | Y_n \cap \hat Y_n \right |}{\left | Y_n \cup \hat Y_n \right |} \nonumber \\
%  \mathcal{P}=\frac{1}{N}\sum_{n=1}^{N}\frac{\left | Y_n \cap \hat Y_n \right |}{\left | \hat Y_n \right |} \nonumber \\
%      \mathcal{R}=\frac{1}{N}\sum_{n=1}^{N}\frac{\left | Y_n \cap \hat Y_n \right |}{\left | Y_n \right |}  \nonumber\\
%      \mathcal{F}=2\left(\frac{\mathcal{P} \times  \mathcal{R}}{\mathcal{P} + \mathcal{R}}\right)
%  \end{gather}

%  where $Y_n$ and $\hat Y_n$ respectively are the ground truth positive labels and predicted positive labels for the $n$-th instance. $|.|$ indicates the set cardinality, and $N$ is the number of instances.

\subsection{Preprocessing}
RAP and PETA samples vary in size, with each image containing exclusively one subject annotated. Therefore, to have constant ratio images, we first performed a zero-padding and then resized them into $256 \times 256$. It worth mentioning that, after each residual block, the input size is divided by 2. Therefore, as we have implemented the backbone with $4$ residual stages, to multiply the binary mask and feature maps with a size of $16 \times 16$, the input size should be $256 \times 256$. Note that the sharp edges caused by these zero pads do not affect the network due to the presence of the \emph{multiplication layer} before the classification layers.

To assure a fair comparison between the tested methods, we used the same train-validation-test splits as in \citep{li2018richly}: $50,957$ images were used for learning, $16,986$ for validation purposes, and the remaining 16,985 images used for testing. The same strategy was used for the PETA dataset. Table \ref{table: Param_Settings} shows the parameter settings of our multi-task network.

\begin{table}[!t]
\scriptsize
\centering
\caption{Parameter Settings for the experiment on RAP dataset.}
\label{table: Param_Settings}
    \begin{tabular}{||m{2.9cm} | M{2cm}||} \hline
       Parameter & Value  \\\hline\hline
        Image input shape & $256 \times 256 \times 3$ \\\hline
        Mask input shape & $16 \times 16 \times 3$ \\\hline
        Learning rate & $1\times e^{-4}$ \\\hline
        Learning decay &  $1\times e^{-6}$ \\\hline
        Number of epochs &  200 \\\hline
        Drop-out probability & 0.7  \\\hline
        Batch size & 8 \\\hline
    \end{tabular}
% \vspace{-0.5 cm}
\end{table}

\subsection{Implementation Details}
Our method was implemented using Keras $2.2.5$ with Tensorflow $1.12.0$ backend \citep{abadi2016tensorflow}, and all the experiments were performed on a machine with an Intel Core i$5-8600$K CPU @ $3.60$ GHz (Hexa Core | $6$ Threads) processor, NVIDIA GeForce RTX $2080$ Ti GPU, and $32$ GB RAM.

 The proposed CNN architecture was fulfilled as a dual-step network. At first, we applied the body segmentation network (\emph{i.e.}, Mask R-CNN, explained in the next subsection) to extract the human full-body masks. We then trained a two-input multi-task network that receives the preprocessed masks and the input data. It is worth mentioning that, on account of the spreading or gathering nature of the attributes features in the full-body human images, we intuitively clustered all the binary attributes into $7$ and $6$ groups for the experiments on RAP and PETA, respectively, as given in Table \ref{table: PETAandRAP_taskpolicy}.

%However, there are some pre-processing steps that are explained in \ref{section: Implementation} subsection.

\begin{table*}[b]
% \vspace{-0.5 cm}
\tiny
\centering
\caption{Task specification policy for PETA and RAP datasets.}
\label{table: PETAandRAP_taskpolicy}
\begin{tabular}{||m{0.7cm}|M{1.9cm}|M{1.2cm}|M{1.8cm}|M{1.7cm}|M{1.6cm}|M{1.8cm}|M{1.8cm}||} \hline
    Dataset & Task 1 (Full Body)  & Task 2 (Head )  & Task 3 (Upper Body)  & Task 4 (Lower Body)  & Task 5 (Foot wears)  & Task 6 (Accessories)  & Task 7 (Action) \\\hline\hline
    PETA  & Female, Male, AgeLess30, AgeLess45, AgeLess60, AgeLarger60 & Hat, LongHair, Scarf, Sunglasses, Nothing   & Casual, Formal, Jacket, Logo, Plaid, ShortSleeves, Strip, Tshirt, Vneck, Other & Casual, Formal, Jeans, Shorts, ShortSkirt, Trousers & LeatherShoes, Sandals, FootwearShoes, Sneaker & Backpack , MessengerBag, PlasticBags, CarryingNothing, CarryingOther & -\\\hline
    RAP & Female, Male, AgeLess16, Age17-30, Age31-45, Age46-60, BodyFat, BodyNormal, BodyThin, Customer, Employee & BaldHead, LongHair, BlackHair, Hat, Glasses & Shirt, Sweater, Vest, TShirt, Cotton, Jacket, SuitUp, Tight, ShortSleeves, Others & LongTrousers, Skirt, ShortSkirt, Dress, Jeans, TightTrousers & Leather, Sports, Boots, Cloth, Casual, Other & Backpack, ShoulderBag, HandBag, Box, PlasticBag, PaperBag, HandTrunk, Other & Calling, Talking,  Gathering, Holding, Pushing, Pulling, CarryingByArm, CarryingByHand \\\hline
\end{tabular}
\end{table*}

As above stated, we used the pre-trained Mask R-CNN \citep{matterport_maskrcnn_2017} to obtain all the foreground masks in our experiments. The used segmentation model was trained in the MS-COCO dataset \citep{lin2014microsoft}. Table \ref{tab:maskrcnn_settings} provides the details of our implementation settings.

% However, as the RPN is adjusted for small aspect ratios and sizes of the anchors, and only is able to process a definite number of proposals (\emph{e.g.,} 256), Mask R-CNN cannot powerfully deal with the large objects in the image. As PAR datasets provide the bounding boxes of the pedestrians, the presented human objects in the image seem massive. There are two solutions to deal with this problem: $i$) Configuring the model parameters according to the expected size of the objects in our data and training the network from scratch. However, in this case, we are not able to use the MS-COCO pre-trained weights provided by \citep{matterport_maskrcnn_2017}. $ii$) Zero padding the images such that desired objects seem non-large from the network points of view. For simplicity, we chose the second solution. Meanwhile, as a result of multiplying the feature maps by a binary mask, the network cannot be distracted by the sharp edges around the image. Having in mind that the images in PAR problems usually cover almost the whole bounding box and can be considered as huge objects, we pre-processed the data and fed the images to the Mask-network with a 200-pixel black border (zero paddings) around human fixed-size patches (\emph{i.e.}, $175 \times 100$).

\begin{table}[!t]
\footnotesize
    \caption{Mask R-CNN parameter settings}
    \label{tab:maskrcnn_settings}
    \centering
    \begin{tabular}{||m{6cm} | M{4cm}||} \hline
         Parameter &  Value\\\hline\hline
         Image input dimension & $1024 \times 1024 \times 3$ \\\hline
         RPN anchor scales  & 32, 64, 128, 256, 512 \\\hline
         RPN anchor ratio & 0.5, 1, 2 \\\hline
         Number of proposals per image & 256 \\\hline
    \end{tabular}
\end{table}

By feeding the input images to the convolutional building blocks, we obtain a set of feature maps that will be multiplied by the corresponding mask, using the element-wise multiplication layer. This layer receives two inputs with the same shapes. Transferring the input data with shape of $256 \times 256 \times 3$ into a $4$-residual block backbone, we obtain a $16 \times 16 \times 1,024$-shaped output. Also, masks are resized to have the same size as the corresponding feature maps. Therefore, as a result of multiplying the binary mask and feature maps, we obtain a set of attention glimpses with the $16 \times 16 \times 1,024$ shape. These \emph{glimpses} are down-sampled to $1,024$ features using a global average pooling layer to decrease the sensitivity of the locations of the features in the input image \citep{lin2013network}. Afterward, in the interest of training one classifier for each task, a $Dense[ReLU]\rightarrow DropOut \rightarrow Dense[ReLU] \rightarrow DropOut \rightarrow Dense[ReLU]\rightarrow Dense[Sigmoid]$ architecture is stacked on top of the shared layers for each task.

% \begin{table}[t]
% \footnotesize
% \centering
% \caption{Desktop computer specifications applied for the experiments}
% \label{table: PC_specifications}
% \begin{tabular}{||m{2.39cm}|M{5cm}||} \hline
%     Specification & Value \\\hline\hline
%     Processor  & Intel Core i5-8600K CPU @ 3.60 GHz (Hexa Core | 6 Threads)\\\hline
%     GPU & NVIDIA GeForce RTX 2080 Ti \\\hline
%     RAM  & 32 GB \\\hline
%     Storage Memory &  620 GB SDD \\ \hline
% \end{tabular}
% \vspace{-0.5 cm}
% \end{table}

%\footnote{Implementation of these methods are available at \url{https://github.com/dangweili/RAP}}

\subsection{Comparison with the State-of-the-art}

\begin{figure}%
\centering
\includegraphics[scale=0.355]{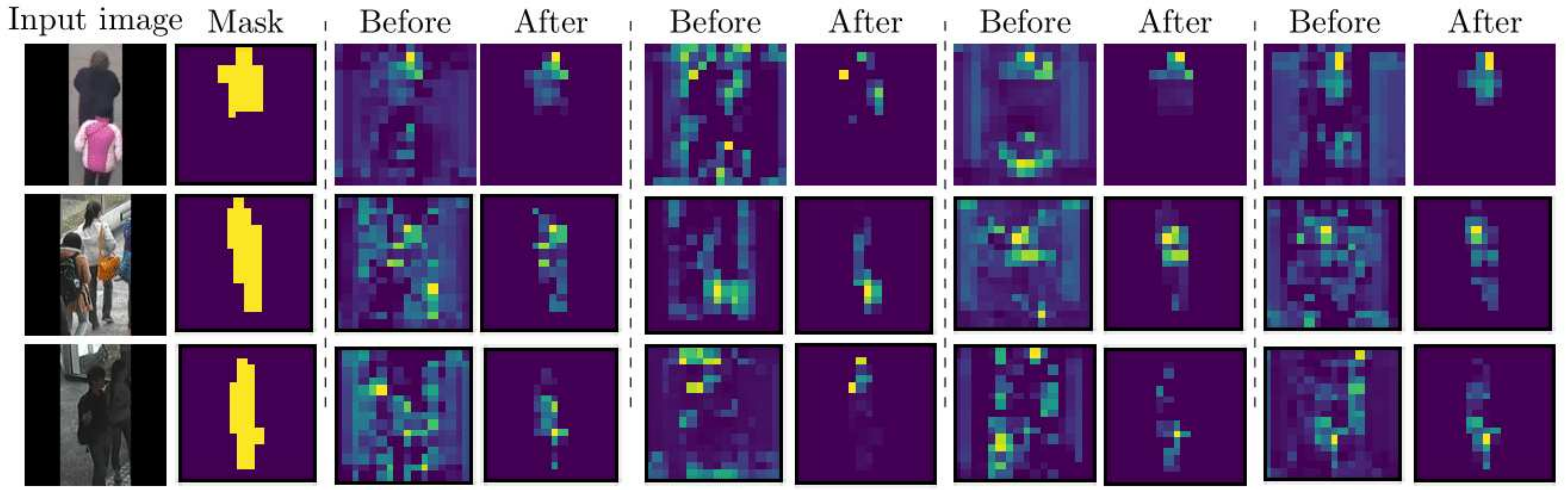}
\caption{The effectiveness of the \emph{multiplication layer} on filtering the background features from the feature maps. The far left column shows the input images to the network, the \emph{Mask} column presents the ground truth binary mask (the first input of the multiplication layer),  the columns with \emph{Before} label (the second input of the multiplication layer) display the feature maps before applying the multiplication operation, and the columns with \emph{After} label show the output of the multiplication layer.}%
\label{fig:feature maps}%
%\vspace{-0.6 cm}
\end{figure}

We compared the performance attained by our method to three baselines, that were considered to represent the state-of-the-art: ACN \citep{sudowe2015person}, DeepMar \citep{li2018richly}, and MLCNN \citep{zhu2017multi} on the RAP and the PETA datasets. These methods have been selected for two reasons: 1- in a way similar to our method, ACN and DeepMar are global-based methods (i.e., they extract features from the full-body images) 2- Authors of these methods have reported the results for all the attributes in a separate way, assuring a fair comparison between the performance of all methods.

As the solution proposed in this paper, the ACN \citep{sudowe2015person} method analyzes the full-body images and jointly learns all the attributes without relying on additional information. DeepMar \citep{li2018richly} is a global-based end-to-end CNN model that provides all the binary labels for the input image, simultaneously. In \citep{zhu2017multi},  authors propose a multi-label convolutional neural network (MLCNN) that divides the input image into  overlapped parts and fuses the features of each CNN to provide the binary labels for the pedestrians. Tables \ref{table: ResultsPETA} and \ref{table: ResultsRAP} provide the obtained results observed for the three methods considered in the PETA and RAP datasets.

% LOMO network \citep{chen2018deep} is also a part-specific CNN that fuses the high-level features of the body parts with low-level local maximal occurrence features and trains a classifier that is able to deal with different body poses and camera views. These methods have achieved comparable results in the PETA dataset. Table \ref{table: ResultsPETA} shows the comparison results between MLCNN, LOMO and our method.

Table \ref{table: ResultsPETA} shows the evaluation results of the DeepMar and MLCNN methods, including our model on the PETA dataset. According to this table, our model shows superior recognition rates for 22 (out of 27) attributes, concluded to more than 3\% improvement in total accuracy. If we consider 35 attributes, the proposed network achieves a 91.7\% recognition rate while this value for the DeepMar approach is 82.6\%.

The experiment carried out without considering image augmentation (\emph{i.e.}, 5-degree rotation, horizontal flip, 0.02 width and height shift range, 0.05 shear range, 0.08 zoom range and changing the brightness in the interval [0.9,1.1]), showed 85.5\% and 88.2\% average accuracy for 27 and 35 attributes, respectively. We augmented the images randomly, and after the visualization of some images, we determined the values in augmentations.

As shown in Table \ref{table: ResultsRAP}, the average recognition rates for the ACN and DeepMar methods respectively were 68.92\% and 75.54\%, while our approach achieved more than 92\%. In particular, excluding five attributes (\emph{i.e.}, \textit{Female}, \textit{Shirt}, \textit{Jacket}, \textit{Long Trousers}, and \textit{Other} class in attachments category), our PAR model provides notoriously better results than the DeepMar method, and better than the ACN model in all cases.

The proposed method shows superior results in both datasets; however, in 22 attributes of the RAP benchmark, the recognition percentage is yet less than 95\%, and in 7 cases, this rate is even less than 80\%. The same interpretation is valid for the PETA dataset as well, which indicates the demands of more research works in the PAR field of study.

\begin{table}[!t]
\scriptsize
\centering
\caption{Comparison between the results observed in the PETA dataset (mean accuracy percentage). The highest accuracy values per attribute among all methods appear in bold.}
\label{table: ResultsPETA}
\begin{tabular}{||m{3cm}|M{2.5cm}|M{2.5cm}|M{1.3cm}||} \hline
 Attributes          & DeepMar \citep{li2018richly} & MLCNN \citep{zhu2017multi}   &   Proposed  \\\hline\hline

 Male                & 89.9 &  84.3   &  \textbf{91.2} \\\cline{1-4}

 AgeLess30           & \textbf{85.8} &  81.1   &  85.3  \\
 AgeLess45           & 81.8 &  79.9   &  \textbf{82.7}  \\
 AgeLess60           & 86.3 &  92.8   &  \textbf{93.9}  \\
 AgeLarger60         & 94.8 &  97.6   &  \textbf{98.6}  \\\hline

 Head-Hat            & 91.8 &  96.1   &  \textbf{97.4}  \\
 Head-LongHair       & 88.9 &  88.1   &  \textbf{92.3}  \\
 Head-Scarf          & 96.1 &  97.2   &  \textbf{98.2}  \\
 Head-Nothing        & 85.8 &  86.1   &  \textbf{90.7}  \\\hline

 UB-Casual           & 84.4 &  89.3   &  \textbf{93.4}  \\
 UB-Formal           & 85.1 &  91.1   &  \textbf{94.6}  \\
 UB-Jacket           & 79.2 &  92.3   &  \textbf{95.0}  \\
 UB-ShortSleeves     & 87.5 &  88.1   &  \textbf{93.4}  \\
 UB-Tshirt           & 83.0 &  90.6   &  \textbf{93.8}  \\
 UB-Other            & \textbf{86.1} &  82.0   &  84.8  \\\hline

 LB-Casual           & 84.9 &  90.5   &  \textbf{93.7}  \\
 LB-Formal           & 85.2 &  90.9   &  \textbf{94.0}  \\
 LB-Jeans            & 85.7 &  83.1   &  \textbf{86.7}  \\
 LB-Trousers         & \textbf{84.3 }&  76.2   &  78.9  \\\hline

 Shoes-Leather       & 87.3 &  85.2   &  \textbf{89.8}  \\
 Shoes-Footwear      & \textbf{80.0} &  75.8   &  79.8  \\
 Shoes-Sneaker       & 78.7 &  81.8   &  \textbf{86.6}  \\\hline

 Backpack            & 82.6 &  84.3   &  \textbf{89.2}  \\
 MessengerBag        & 82.0 &  79.6   &  \textbf{86.3}  \\
 PlasticBags         & 87.0 &  93.5   &  \textbf{94.5}  \\
 Carrying-Nothing    & 83.1 &  80.1   &  \textbf{85.9}  \\
 Carrying-Other      & 77.3 &  \textbf{80.9}   &  78.8  \\\hline\hline

 Average of 27 Att.  & 85.4 &  86.6   &  \textbf{90.0} \\\hline
 Average of 35 Att.  & 82.6 &   -     &  \textbf{91.7} \\\hline
\end{tabular}
% \vspace{-0.6 cm}
\end{table}

\begin{table}[!t]
\scriptsize
\centering
\caption{Comparison of the results observed in the RAP dataset (mean accuracy percentage). The highest accuracy values per attribute among all methods appear in bold.}
\label{table: ResultsRAP}
\begin{tabular}{||m{3cm}|M{2cm}|M{2cm}|M{2cm}||} \hline
  Attributes & ACN \citep{sudowe2015person} & DeepMar \citep{li2018richly} & Proposed  \\\hline\hline

    Female           & 94.06 & \textbf{96.53} & 96.28 \\\hline

    AgeLess16        & 77.29 & 77.24 & \textbf{99.25} \\
    Age17-30         & 69.18 & 69.66 & \textbf{69.98} \\
    Age31-45         & 66.80 & 66.64 & \textbf{67.19} \\
    Age46-60         & 52.16 & 59.90 & \textbf{96.88} \\\hline

    BodyFat          & 58.42 & 61.95 & \textbf{87.24} \\
    BodyNormal       & 55.36 & 58.47 & \textbf{78.20} \\
    BodyThin         & 52.31 & 55.75 & \textbf{92.82} \\\cline{1-4}

    Customer         & 80.85 & 82.30 & \textbf{96.98} \\
    Employee         & 85.60 & 85.73 & \textbf{97.67} \\\cline{1-4}

    BaldHead         & 65.28 & 80.93 & \textbf{99.56}  \\
    LongHair         & 89.49 & 92.47 & \textbf{94.67} \\
    BlackHair        & 66.19 & 79.33 & \textbf{94.94} \\\cline{1-4}

    Hat              & 60.73 & 84.00 & \textbf{99.02} \\
    Glasses          & 56.30 & 84.19 & \textbf{96.76} \\\hline

    UB-Shirt            & 81.81 & \textbf{85.86} & 83.93  \\
    UB-Sweater          & 56.85 & 64.21 & \textbf{92.66}  \\
    UB-Vest             & 83.65 & 89.91 & \textbf{96.91}  \\
    UB-TShirt           & 71.61 & 75.94 & \textbf{77.17} \\
    UB-Cotton           & 74.67 & 79.02 & \textbf{89.48} \\
    UB-Jacket           & 78.29 & \textbf{80.69} & 71.93 \\
    UB-SuitUp           & 73.92 & 77.29 & \textbf{97.18} \\
    UB-Tight            & 61.71 & 68.89 & \textbf{96.10} \\
    UB-ShortSleeves      & 88.27 & 90.09 & \textbf{90.79} \\
    UB-Others           & 50.35 & 54.82 & \textbf{97.91} \\\cline{1-4}

    LB-LongTrousers     & 86.60 & \textbf{86.64} & 84.88  \\
    LB-Skirt            & 70.51 & 74.83 & \textbf{97.37}  \\
    LB-ShortSkirt       & 73.16 & 72.86 & \textbf{98.10} \\
    LB-Dress            & 72.89 & 76.30 & \textbf{97.34} \\
    LB-Jeans            & 90.17 & 89.46 & \textbf{91.56} \\
    LB-TightTrousers    & 86.95 & 87.91 & \textbf{94.71} \\\cline{1-4}

    shoes-Leather    & 71.92 & 80.50 & \textbf{84.00}  \\
    shoes-Sports     & 62.59 & 71.58 & \textbf{80.68}  \\
    shoes-Boots      & 85.03 & 91.37 & \textbf{96.68}  \\
    shoes-Cloth      & 68.74 & 72.31 & \textbf{98.67}  \\
    shoes-Casual     & 54.57 & 64.58 & \textbf{77.74} \\
    shoes-Other      & 52.42 & 61.56 & \textbf{92.00} \\\cline{1-4}

    Backpack         & 68.87 & 80.61 & \textbf{98.03} \\
    ShoulderBag      & 69.30 & 82.52 & \textbf{93.29}  \\
    HandBag          & 63.95 & 76.45 & \textbf{97.64}  \\
    Box              & 66.72 & 76.18 & \textbf{96.30}  \\
    PlasticBag       & 61.53 & 75.20 & \textbf{97.78} \\
    PaperBag         & 52.25 & 63.34 & \textbf{99.07} \\
    HandTrunk        & 79.01 & 84.57 & \textbf{97.74}  \\
    Other            & 66.14 & \textbf{76.14} & 71.54 \\\cline{1-4}

    Calling          & 74.66 & 86.97 & \textbf{97.13}  \\
    Talking          & 50.54 & 54.65 & \textbf{97.54} \\
    Gathering        & 52.69 & 58.81 & \textbf{95.47}  \\
    Holding          & 56.43 & 64.22 & \textbf{97.71} \\
    Pushing          & 80.97 & 82.58 & \textbf{99.15} \\
    Pulling          & 69.00 & 78.35 & \textbf{98.24} \\
    CarryingByArm    & 53.55 & 65.40 & \textbf{97.77}  \\
    CarryingByHand   & 74.58 & 82.72 & \textbf{87.57}  \\
    Other            & 54.83 & 58.79 & \textbf{99.13}  \\\hline\hline
    Average          & 68.92 & 75.54 & \textbf{92.23}  \\\hline
\end{tabular}
% \vspace{-0.5 cm}
\end{table}

\subsection{Ablation Studies}
In this section, we study the effectiveness of the mentioned contributions in Fig. \ref{fig: Network}. To this end, we trained and tested a light version of the network (with three residual blocks and input image size $128 \times 128$) on the PETA dataset with similar initialization, but different settings (Table \ref{table: ablation-study}). The first row of Table \ref{table: ablation-study} shows the performance of a network, constructed from three residual blocks with four shared fully connected layers on top, plus one fully connected layer for each attribute. In this architecture, as the system cannot decide on each task independently, the performance is poor ($81.11 \%$), and the network cannot predict the uncorrelated attributes (e.g., behavioral attributes versus appearance attributes) effectively. However, the results in the second row of Table \ref{table: ablation-study} show that repeating the fully connected layers for each task independently (while keeping the rest of the architecture unchanged), improves the results by around $8 \%$. Furthermore, equipping the network with the proposed weighted loss function (Table \ref{table: ablation-study}, row $3$) and adding the \emph{Multiplication layer} (Table \ref{table: ablation-study}, row $4$) showed further improvements in the performance to $89.35 \%$ and $89.73 \%$, respectively.

\begin{table}[!t]
\scriptsize
\centering
\caption {Ablation studies. The first row shows our baseline system with a multi-label architecture and binary-cross-entropy loss function, while the other rows indicate the proposed system with various settings.}
\label{table: ablation-study}
    \begin{tabular}{|| M{3cm} | M{3cm} | M{3cm} | M{2cm}||} \hline
  Multi-task architecture & Multiplication Layer & Weighted Loss (Binary-cross-entropy)  & mAP (\%)  \\\hline\hline
      -      &  -      &  -      & 81.11 \\\hline
      \cmark &  -      &  -      & 89.18 \\\hline
      \cmark &  -      &  \cmark & 89.35 \\\hline
      \cmark &  \cmark &  -      & 89.73 \\\hline
    \end{tabular}
% \vspace{-0.5 cm}
\end{table}

\textbf{Feature map visualization.} Neural networks are known as poorly interpretable models. However, as the internal structures of the CNNs are designed to operate upon two-dimensional images, they preserve the spatial relationships for what it is being learned \cite{goodfellow2016deep}. Hence, by visualizing the operations on each layer, we can understand the behavior of the network. As a result of slicing the small linear filters over the input data, we obtain the activation maps (feature maps). To analyze the behavior of the proposed \emph{multiplication layer} (Fig. \ref{fig: Network}), we visualized the input and output feature maps in Fig. \ref{fig:feature maps}, such that the columns labeled as \emph{Mask} and \emph{Before} refer to the inputs of the layer, and the columns labeled as \emph{After} show the multiplication results of the two inputs. As it is evident, unwanted features resulting from the partial occlusions were filtered from the feature map, which improved the overall performance of the system.

\textbf{Where is the network looking at?} As a general behavior, CNNs infer what could be the optimal local/global features of a training set and generalize them to decide on unseen data. Here, partial occlusions can easily affect this behavior and decrease the performance, being helpful to understand where the model is actually looking at in the prediction phase. To this end, we plot some heat maps to investigate the effectiveness of the proposed \emph{multiplication layer} and \emph{task-oriented architecture}. Heat maps are easily understandable and highlight the regions on which the network focuses while making a prediction.

\begin{figure}[t!]
\centering
\includegraphics[scale=0.36]{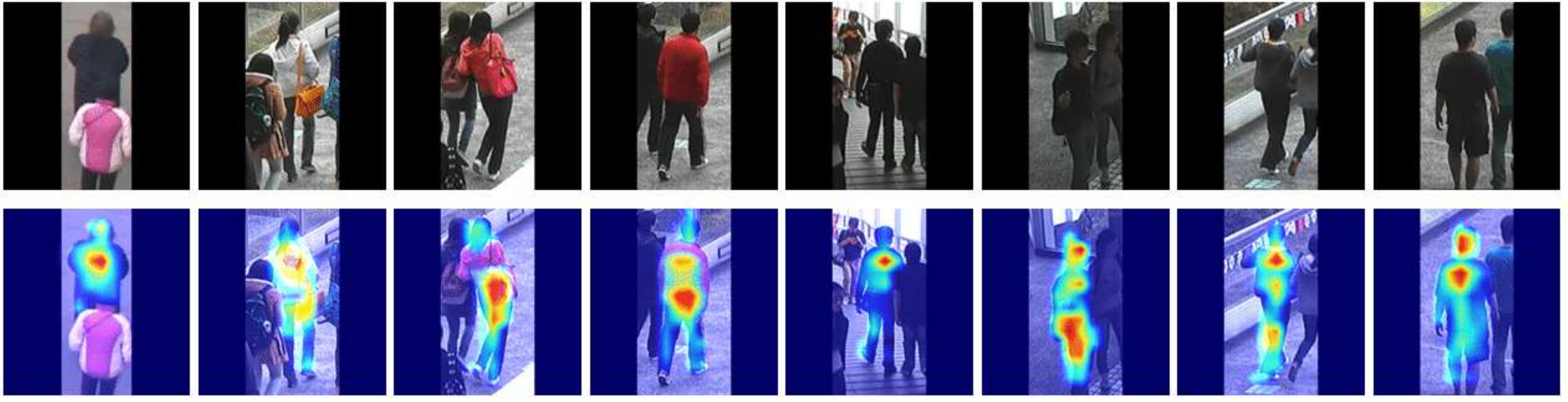}
\caption{Illustration of the effectiveness of the \emph{multiplication layer} upon the focus ability of the proposed model in case of partial occlusions. Samples regard the PETA dataset, with the network predicting the \emph{age} and \emph{gender} attributes.}%
\label{fig:masking_heatmap}%
%\vspace{-0.6 cm}
\end{figure}

\begin{figure}[t!]
\centering
\includegraphics[scale=0.4]{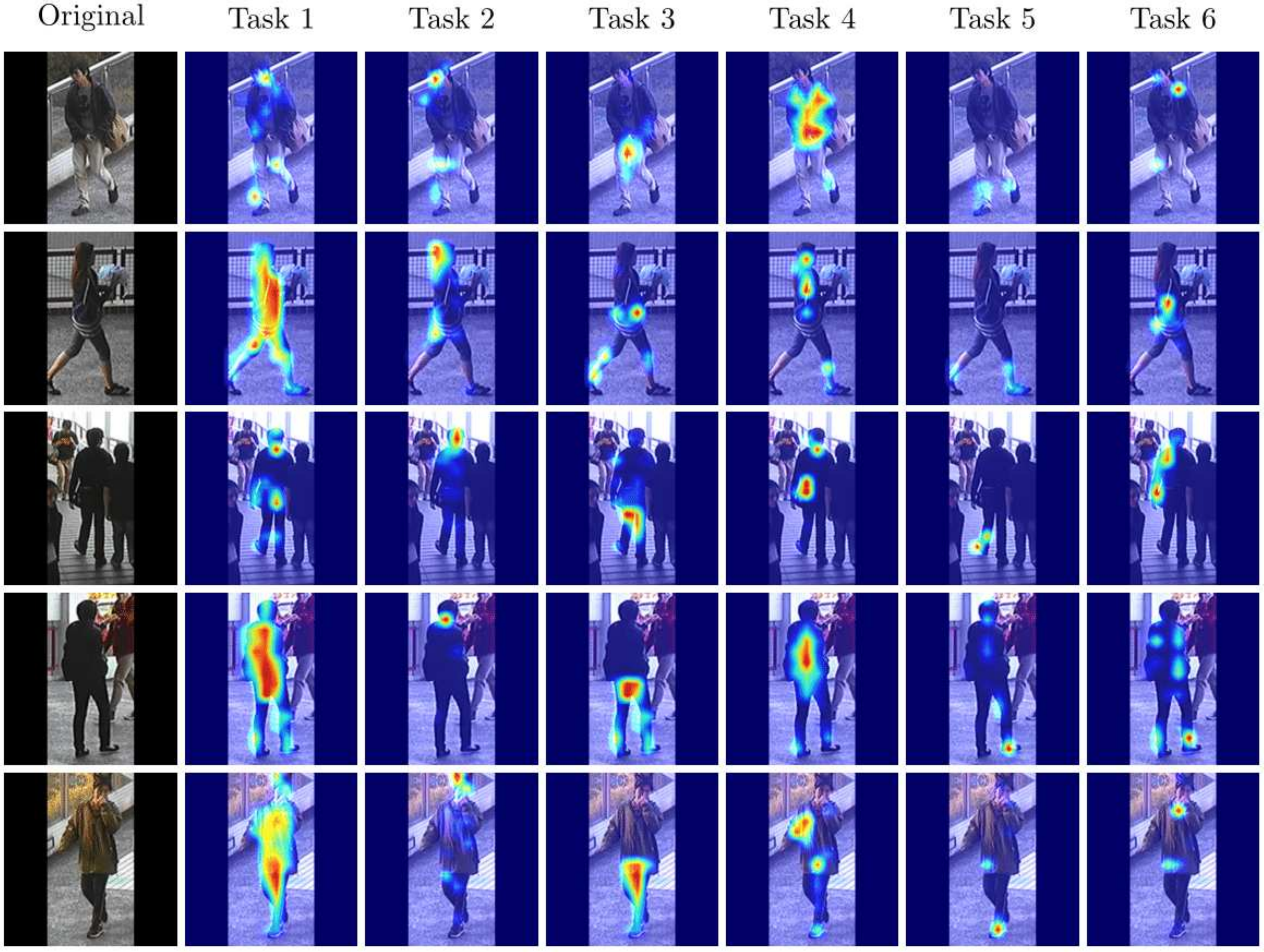}
\caption{Visualization of the heat maps resulting of the proposed multi-task network. Sample regard the PETA dataset. The leftmost column shows the original samples, the column \emph{Task 1} (i.e., recognizing \emph{age} and \emph{gender}) presents the effectiveness of the network focus on the human full-body, and the remaining columns display the ability of the system on region-based attribute recognition. The task policies are given in Table \ref{table: PETAandRAP_taskpolicy}.}
\label{fig:body_part_heatmap}%
% \vspace{-0.6 cm}
\end{figure}

Fig. \ref{fig:masking_heatmap} shows the behavior of the system regarding the examples with partial occlusions. As it is seen, the proposed network is able to filter the harmful features of the distractors effectively, while focusing on the target subject. Moreover, Fig. \ref{fig:body_part_heatmap} shows the model behavior during the attribute recognition in each task.

\textbf{Loss Function.}
Table \ref{table: loss_function} provides the performance of the proposed network, when using different loss functions suitable for binary classification. Focal loss \cite{lin2017focal} forces the network to concentrate on hard samples, while the weighted Binary Cross-Entropy (BCE) loss \cite{li2015multi} allocates a specific binary weight to each class. Training the network using binary focal loss function showed $79.30 \%$ accuracy in the test phase, while this number was $90.19 \%$ for the weighted BCE loss (see Table \ref{table: loss_function}).

The proposed weighted loss function uses the BCE loss function, while recommends different weights for each \emph{class}. We further trained the proposed model with the binary focal loss function using the proposed weights. The results in Table \ref{table: loss_function} indicate a slight improvement in the performance when we train the network using the proposed weighted loss function with BCE ($90.34 \%$).

%In sample-based loss, the number of samples in each class determines the weight, while in task-based loss function, we have given an equal importance to each task.

\begin{table}[!t]
\scriptsize
\centering
\caption{Performance of the network trained with different loss functions on PETA dataset.}
\label{table: loss_function}
    \begin{tabular}{||m{8cm} | M{2cm}||} \hline
       Loss Function & mAP (\%)  \\\hline\hline
        Binary focal loss function \cite{lin2017focal} & 79.30 \\\hline
        Weighted BCE loss function \cite{li2015multi} & 90.19 \\\hline
        %Sample Based Loss & 90.33 \\\hline
        Proposed weighted loss function (with BCE) &  90.34 \\\hline
        Proposed weighted loss function (with binary focal loss) &  89.27 \\\hline
    \end{tabular}
% \vspace{-0.5 cm}
\end{table}

\section{Conclusions}
\label{sec: Conclusion}
Complex background clutter, viewpoint variations, and occlusions are known to have a noticeable negative effect on the performance of person attribute recognition (PAR) methods. According to this observation, in this paper, we proposed a deep-learning framework that improves the robustness of the obtained feature representation by directly discarding the background regions in the fully connected layers of the network. To this end, we described an element-wise multiplication layer between the output of the residual convolutional layers and a binary mask representing the human full-body foreground. Further, the refined feature maps were down-sampled and fed to different fully connected layers, that each one is specialized in learning a particular task (i.e., a subset of attributes). Finally, we described a loss function that weights each category of attributes to ensure that each attribute receives enough attention, and there are not some attributes that bias the results of others. Our experimental analysis on the PETA and RAP datasets pointed for solid improvements in the performance of the proposed model with respect to the state-of-the-art.

\section*{Acknowledgments}
This research is funded by the “FEDER, Fundo de Coesao e Fundo Social Europeu” under the “PT2020 - Portugal 2020” program, “IT: Instituto de Telecomunicações” and “TOMI: City’s Best Friend.” Also, the work is funded by FCT/MEC through national funds and, when applicable, co-funded by the FEDER PT2020 partnership agreement under the project UID/EEA/50008/2019.

%\bibliographystyle{model2-names}
%\bibliography{refs}

%% References


\bibliographystyle{elsarticle-num}

\bibliography{Main}

\begin{thebibliography}{10}
\expandafter\ifx\csname url\endcsname\relax
  \def\url#1{\texttt{#1}}\fi
\expandafter\ifx\csname urlprefix\endcsname\relax\def\urlprefix{URL }\fi
\expandafter\ifx\csname href\endcsname\relax
  \def\href#1#2{#2} \def\path#1{#1}\fi

\bibitem{mabrouk2018abnormal}
A.~B. Mabrouk, E.~Zagrouba, Abnormal behavior recognition for intelligent video
  surveillance systems: A review, Expert Systems with Applications 91 (2018)
  480--491.

\bibitem{kumari2015facial}
J.~Kumari, R.~Rajesh, K.~Pooja, Facial expression recognition: A survey,
  Procedia Computer Science 58 (2015) 486--491.

\bibitem{schmidhuber2015deep}
J.~Schmidhuber, Deep learning in neural networks: An overview, Neural networks
  61 (2015) 85--117.

\bibitem{liu2017survey}
W.~Liu, Z.~Wang, X.~Liu, N.~Zeng, Y.~Liu, F.~E. Alsaadi, A survey of deep
  neural network architectures and their applications, Neurocomputing 234
  (2017) 11--26.

\bibitem{wang2019pedestrian}
X.~Wang, S.~Zheng, R.~Yang, B.~Luo, J.~Tang, Pedestrian attribute recognition:
  A survey, arXiv preprint arXiv:1901.07474.

\bibitem{li2015multi}
D.~Li, X.~Chen, K.~Huang, Multi-attribute learning for pedestrian attribute
  recognition in surveillance scenarios, in: 2015 3rd IAPR Asian Conference on
  Pattern Recognition (ACPR), IEEE, 2015, pp. 111--115.

\bibitem{sudowe2015person}
P.~Sudowe, H.~Spitzer, B.~Leibe, Person attribute recognition with a
  jointly-trained holistic cnn model, in: Proceedings of the IEEE International
  Conference on Computer Vision Workshops, 2015, pp. 87--95.

\bibitem{abdulnabi2015multi}
A.~H. Abdulnabi, G.~Wang, J.~Lu, K.~Jia, Multi-task cnn model for attribute
  prediction, IEEE Transactions on Multimedia 17~(11) (2015) 1949--1959.

\bibitem{liu2018localization}
P.~Liu, X.~Liu, J.~Yan, J.~Shao, Localization guided learning for pedestrian
  attribute recognition, arXiv preprint arXiv:1808.09102.

\bibitem{gkioxari2015actions}
G.~Gkioxari, R.~Girshick, J.~Malik, Actions and attributes from wholes and
  parts, in: Proceedings of the IEEE International Conference on Computer
  Vision, 2015, pp. 2470--2478.

\bibitem{li2016human}
Y.~Li, C.~Huang, C.~C. Loy, X.~Tang, Human attribute recognition by deep
  hierarchical contexts, in: European Conference on Computer Vision, Springer,
  2016, pp. 684--700.

\bibitem{chen2018pedestrian}
Y.~Chen, S.~Duffner, A.~STOIAN, J.-Y. Dufour, A.~Baskurt,
  \href{https://hal.archives-ouvertes.fr/hal-01625470}{{Pedestrian attribute
  recognition with part-based CNN and combined feature representations}}, in:
  {VISAPP2018}, Funchal, Portugal, 2018.
\newline\urlprefix\url{https://hal.archives-ouvertes.fr/hal-01625470}

\bibitem{sarafianos2018deep}
N.~Sarafianos, X.~Xu, I.~A. Kakadiaris, Deep imbalanced attribute
  classification using visual attention aggregation, in: Proceedings of the
  European Conference on Computer Vision (ECCV), 2018, pp. 680--697.

\bibitem{sarfraz2017deep}
M.~S. Sarfraz, A.~Schumann, Y.~Wang, R.~Stiefelhagen, Deep view-sensitive
  pedestrian attribute inference in an end-to-end model, arXiv preprint
  arXiv:1707.06089.

\bibitem{li2018richly}
D.~Li, Z.~Zhang, X.~Chen, K.~Huang, A richly annotated pedestrian dataset for
  person retrieval in real surveillance scenarios, IEEE transactions on image
  processing 28~(4) (2018) 1575--1590.

\bibitem{zhu2017multi}
J.~Zhu, S.~Liao, Z.~Lei, S.~Z. Li, Multi-label convolutional neural network
  based pedestrian attribute classification, Image and Vision Computing 58
  (2017) 224--229.

\bibitem{PARatFAR}
Y.~Deng, P.~Luo, C.~C. Loy, X.~Tang,
  \href{http://doi.acm.org/10.1145/2647868.2654966}{Pedestrian attribute
  recognition at far distance}, in: Proceedings of the 22Nd ACM International
  Conference on Multimedia, MM '14, ACM, New York, NY, USA, 2014, pp. 789--792.
\newblock \href {http://dx.doi.org/10.1145/2647868.2654966}
  {\path{doi:10.1145/2647868.2654966}}.
\newline\urlprefix\url{http://doi.acm.org/10.1145/2647868.2654966}

\bibitem{zhu2013pedestrian}
J.~Zhu, S.~Liao, Z.~Lei, D.~Yi, S.~Li, Pedestrian attribute classification in
  surveillance: Database and evaluation, in: Proceedings of the IEEE
  international conference on computer vision workshops, 2013, pp. 331--338.

\bibitem{layne2014attributes}
R.~Layne, T.~M. Hospedales, S.~Gong, Attributes-based re-identification, in:
  Person Re-Identification, Springer, 2014, pp. 93--117.

\bibitem{tan2019attention}
Z.~Tan, Y.~Yang, J.~Wan, H.~Wan, G.~Guo, S.~Z. Li, Attention based pedestrian
  attribute analysis, IEEE transactions on image processing.

\bibitem{li2019visual}
Q.~Li, X.~Zhao, R.~He, K.~Huang, Visual-semantic graph reasoning for pedestrian
  attribute recognition, in: Proceedings of the AAAI Conference on Artificial
  Intelligence, Vol.~33, 2019, pp. 8634--8641.

\bibitem{zhao2019recurrent}
X.~Zhao, L.~Sang, G.~Ding, J.~Han, N.~Di, C.~Yan, Recurrent attention model for
  pedestrian attribute recognition, in: Proceedings of the AAAI Conference on
  Artificial Intelligence, Vol.~33, 2019, pp. 9275--9282.

\bibitem{lou2019mse}
M.~Lou, Z.~Yu, F.~Guo, X.~Zheng, Mse-net: Pedestrian attribute recognition
  using mlsc and se-blocks, in: International Conference on Artificial
  Intelligence and Security, Springer, 2019, pp. 217--226.

\bibitem{zhu2015multi}
J.~Zhu, S.~Liao, D.~Yi, Z.~Lei, S.~Z. Li, Multi-label cnn based pedestrian
  attribute learning for soft biometrics, in: 2015 International Conference on
  Biometrics (ICB), IEEE, 2015, pp. 535--540.

\bibitem{wang2017attribute}
J.~Wang, X.~Zhu, S.~Gong, W.~Li, Attribute recognition by joint recurrent
  learning of context and correlation, in: Proceedings of the IEEE
  International Conference on Computer Vision, 2017, pp. 531--540.

\bibitem{li2018pose}
D.~Li, X.~Chen, Z.~Zhang, K.~Huang, Pose guided deep model for pedestrian
  attribute recognition in surveillance scenarios, in: 2018 IEEE International
  Conference on Multimedia and Expo (ICME), IEEE, 2018, pp. 1--6.

\bibitem{yang2016attribute}
L.~Yang, L.~Zhu, Y.~Wei, S.~Liang, P.~Tan, Attribute recognition from adaptive
  parts, arXiv preprint arXiv:1607.01437.

\bibitem{Liu_2017_ICCV}
X.~Liu, H.~Zhao, M.~Tian, L.~Sheng, J.~Shao, S.~Yi, J.~Yan, X.~Wang,
  Hydraplus-net: Attentive deep features for pedestrian analysis, in:
  Proceedings of the IEEE international conference on computer vision, 2017,
  pp. 350--359.

\bibitem{he2016deep}
K.~He, X.~Zhang, S.~Ren, J.~Sun, Deep residual learning for image recognition,
  in: Proceedings of the IEEE conference on computer vision and pattern
  recognition, 2016, pp. 770--778.

\bibitem{he2016identity}
K.~He, X.~Zhang, S.~Ren, J.~Sun, Identity mappings in deep residual networks,
  in: European conference on computer vision, Springer, 2016, pp. 630--645.

\bibitem{he2017mask}
K.~He, G.~Gkioxari, P.~Doll{\'a}r, R.~Girshick, Mask r-cnn, in: Proceedings of
  the IEEE international conference on computer vision, 2017, pp. 2961--2969.

\bibitem{ren2015faster}
S.~Ren, K.~He, R.~Girshick, J.~Sun, Faster r-cnn: Towards real-time object
  detection with region proposal networks, in: Advances in neural information
  processing systems, 2015, pp. 91--99.

\bibitem{long2015fully}
J.~Long, E.~Shelhamer, T.~Darrell, Fully convolutional networks for semantic
  segmentation, in: Proceedings of the IEEE conference on computer vision and
  pattern recognition, 2015, pp. 3431--3440.

\bibitem{DBLP}
S.~Ruder, \href{http://arxiv.org/abs/1706.05098}{An overview of multi-task
  learning in deep neural networks}, CoRR abs/1706.05098.
\newblock \href {http://arxiv.org/abs/1706.05098} {\path{arXiv:1706.05098}}.
\newline\urlprefix\url{http://arxiv.org/abs/1706.05098}

\bibitem{liu2015deep}
Z.~Liu, P.~Luo, X.~Wang, X.~Tang, Deep learning face attributes in the wild,
  in: Proceedings of the IEEE international conference on computer vision,
  2015, pp. 3730--3738.

\bibitem{he2017adaptively}
K.~He, Z.~Wang, Y.~Fu, R.~Feng, Y.-G. Jiang, X.~Xue, Adaptively weighted
  multi-task deep network for person attribute classification, in: Proceedings
  of the 25th ACM international conference on Multimedia, ACM, 2017, pp.
  1636--1644.

\bibitem{lin2019improving}
Y.~Lin, L.~Zheng, Z.~Zheng, Y.~Wu, Z.~Hu, C.~Yan, Y.~Yang, Improving person
  re-identification by attribute and identity learning, Pattern Recognition.

\bibitem{abadi2016tensorflow}
M.~Abadi, P.~Barham, J.~Chen, Z.~Chen, A.~Davis, J.~Dean, M.~Devin,
  S.~Ghemawat, G.~Irving, M.~Isard, et~al., Tensorflow: A system for
  large-scale machine learning, in: 12th $\{$USENIX$\}$ Symposium on Operating
  Systems Design and Implementation ($\{$OSDI$\}$ 16), 2016, pp. 265--283.

\bibitem{matterport_maskrcnn_2017}
W.~Abdulla, Mask r-cnn for object detection and instance segmentation on keras
  and tensorflow, \url{https://github.com/matterport/Mask_RCNN} (2017).

\bibitem{lin2014microsoft}
T.-Y. Lin, M.~Maire, S.~Belongie, J.~Hays, P.~Perona, D.~Ramanan,
  P.~Doll{\'a}r, C.~L. Zitnick, Microsoft coco: Common objects in context, in:
  European conference on computer vision, Springer, 2014, pp. 740--755.

\bibitem{lin2013network}
M.~Lin, Q.~Chen, S.~Yan, Network in network, arXiv preprint arXiv:1312.4400.

\bibitem{goodfellow2016deep}
I.~Goodfellow, Y.~Bengio, A.~Courville, Deep learning, MIT press, 2016.

\bibitem{lin2017focal}
T.-Y. Lin, P.~Goyal, R.~Girshick, K.~He, P.~Doll{\'a}r, Focal loss for dense
  object detection, in: Proceedings of the IEEE international conference on
  computer vision, 2017, pp. 2980--2988.

\end{thebibliography}

\end{document}